\pdfoutput=1
\documentclass[runningheads]{llncs}
\usepackage{subfig}
\usepackage{caption}
\usepackage{bm}
\usepackage{graphicx}
\usepackage{indentfirst}
\usepackage{amsmath, hyperref}
\usepackage{booktabs,multirow,threeparttable}
\usepackage{colortbl}
\usepackage[export]{adjustbox}
\usepackage{url}
\usepackage{marvosym}
\titlerunning{EdgeNeRF}
\author{Weiqi Yu \and
        Yiyang Yao \and
        Lin He \and 
        Jianming Lv\textsuperscript{(\Letter)}}
\institute{
    School of Computer Science and Technology,\\
    South China University of Technology,\\
    Guangzhou, China\\
    \email{yu\_weiqi@qq.com},
    \email{yiyang.yao.scut@gmail.com},
    \email{481761505@qq.com},
    \email{jmlv@scut.edu.cn}
}
\begin{document}
\definecolor{myred}{RGB}{255,190,190}
\definecolor{myorange}{RGB}{255,225,190}
\definecolor{myyellow}{RGB}{255,245,210}

\title{EdgeNeRF: Edge-Guided Regularization for Neural Radiance Fields from Sparse Views}
%
%\titlerunning{Abbreviated paper title}
% If the paper title is too long for the running head, you can set
% an abbreviated paper title here
%

%
\maketitle              % typeset the header of the contribution
\begin{abstract}
Neural Radiance Fields (NeRF) achieve remarkable performance in dense multi-view scenarios, but their reconstruction quality degrades significantly under sparse inputs due to geometric artifacts. Existing methods utilize global depth regularization to mitigate artifacts, leading to the loss of geometric boundary details. To address this problem, we propose EdgeNeRF, an edge-guided sparse-view 3D reconstruction algorithm. Our method leverages the prior that abrupt changes in depth and normals generate edges. Specifically, we first extract edges from input images, then apply depth and normal regularization constraints to non-edge regions, enhancing geometric consistency while preserving high-frequency details at boundaries. Experiments on LLFF and DTU datasets demonstrate EdgeNeRF’s superior performance, particularly in retaining sharp geometric boundaries and suppressing artifacts. Additionally, the proposed edge-guided depth regularization module can be seamlessly integrated into other methods in a plug-and-play manner, significantly improving their performance without substantially increasing training time. Code is available at \url{https://github.com/skyhigh404/edgenerf}.

\keywords{Neural Radiance Fields \and 3D Reconstruction \and  Sparse Views.}
\end{abstract}

\begin{figure}
\centering	
\subfloat[RegNeRF]{\includegraphics[width=0.49\textwidth]{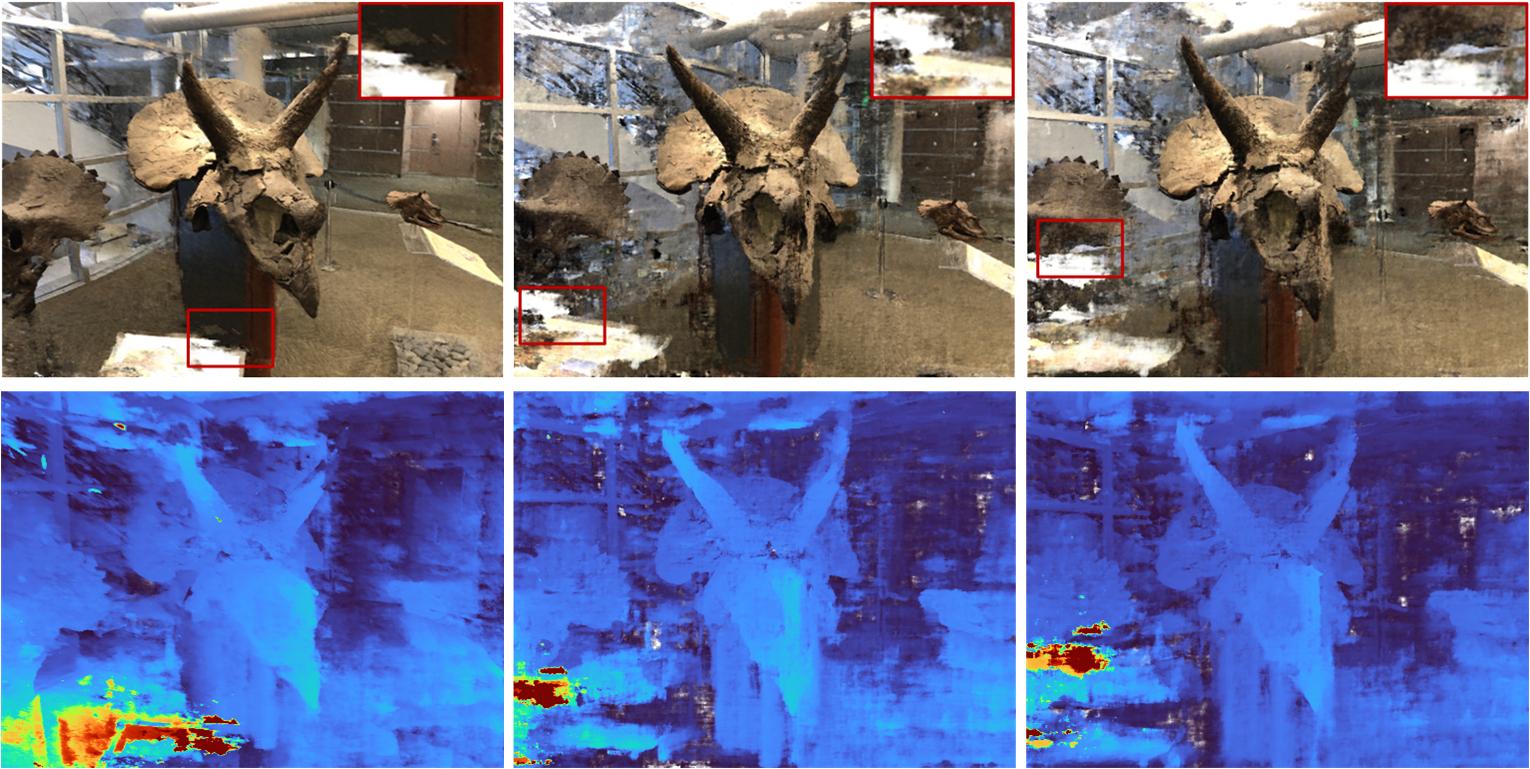}}%
\hfill
\subfloat[EdgeNeRF (ours)]{\includegraphics[width=0.49\textwidth]{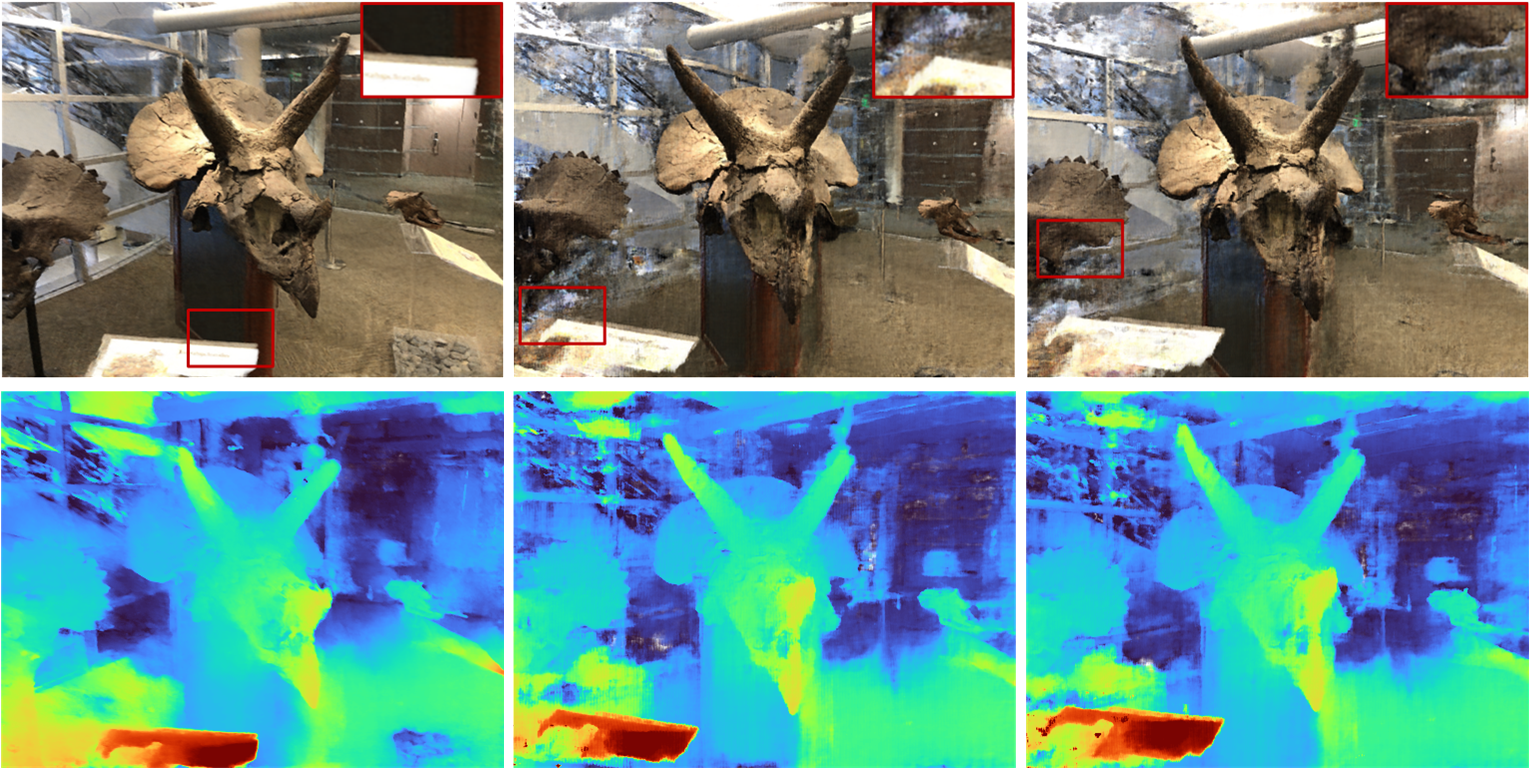}}
\\
\subfloat[Edge Images]{\includegraphics[width=0.49\textwidth]{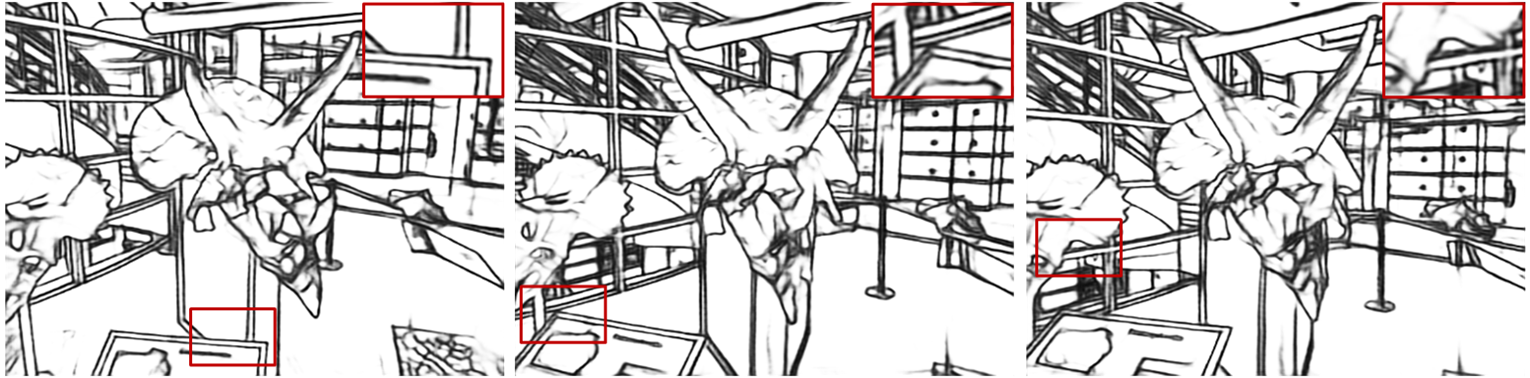}}%
\hfill
\subfloat[Novel Views (GT)]{\includegraphics[width=0.49\textwidth]{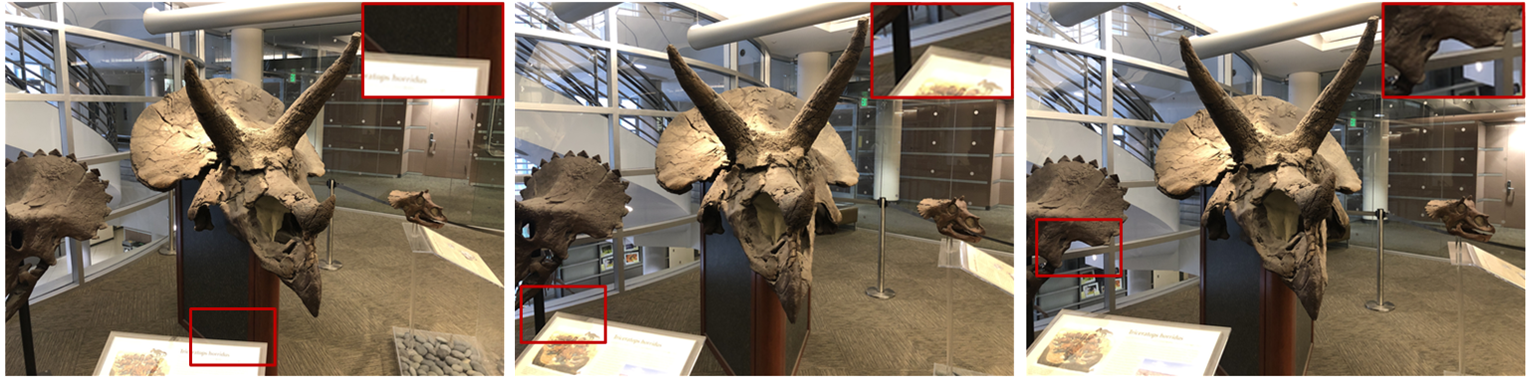}}
\caption{Example of novel view synthesis from sparse views. Our edge-guided regularization yields sharper boundaries and more consistent geometry than state-of-the-art RegNeRF\cite{niemeyer2021regnerfregularizingneuralradiance}.}
\vspace{-12pt}
\label{fig:regnerf_example}
\end{figure}

\section{Introduction}

Neural Radiance Fields (NeRF)\cite{mildenhall2020nerfrepresentingscenesneural} have demonstrated remarkable performance in modeling complex scenes by learning continuous volumetric radiance fields through coordinate-based multilayer perceptrons (MLPs). This breakthro-\\ugh has emerged as a transformative technology with versatile applications across multiple domains, powering 3D digital asset scanning\cite{balloni2023few} and content generation \cite{Latent-NeRF,RODIN} in computer graphics, enabling large-scale urban scene reconstruc-\\tion\cite{Block-nerf,Mega-nerf}. However, NeRF typically requires hundreds of input images from diverse viewpoints, which is an impractical assumption in real-world scenarios due to limited equipment, high acquisition costs, and restricted environments accessibility. Under sparse-view conditions, NeRF’s implicit representation is prone to converging to local optima, resulting in noticeable artifacts and compromised geometric consistency, which negatively impacts both novel view synthesis and subsequent tasks.

Recent advances in sparse-view 3D reconstruction primarily follow two parad-\\igms: pretraining-dependent approaches and regulari\-zation-driven optimization. The former\cite{chen2021mvsnerffastgeneralizableradiance,chibane2021stereoradiancefieldssrf,yu2021pixelnerfneuralradiancefields} learn cross-scene priors through pre-training on large
multi-scene datasets, enabling fast inference through learned feature representations. While effective, these methods require expensive data collection and struggle with domain generalization. On the other hand, regularization-driven methods\\ \cite{kim2022infonerfrayentropyminimization,jain2021puttingnerfdietsemantically,niemeyer2021regnerfregularizingneuralradiance} circumvent pre-training by introducing scene-specific constraints during per-scene optimization. While partially mitigate artifacts, their global regularization strategies often smooth out fine details at object boundaries.

To this end, we propose \textbf{EdgeNeRF}, an edge-guided sparse-view 3D reconstruction algorithm. Our method leverages the observation that abrupt changes in depth or normals often correspond to edges of image. Specifically, we first extract edges from input images, then apply depth and normal regularization constraints to non-edge regions, enhancing geometric consistency while preserving high-frequency details at boundaries, as shown in Figure \ref{fig:regnerf_example}.

Our contributions are summarized as follows:
\begin{enumerate}
\item We propose an explicit edge-guided geometric consistency constraint for sparse-view reconstruction. By encouraging smooth depth and normal variations in non-edge regions while preserving natural discontinuities at edges, our method overcomes the over-smoothing issue common in prior global regularization approaches, significantly improving reconstruction quality and geometric consistency.
\item Our edge-guided depth regularization module is modular and lightweight, making it easily pluggable into existing frameworks to boost performance without introducing significant computational overhead.
\item Experiments on standard sparse view benchmarks demonstrate the superiority of EdgeNeRF. Our method achieves a PSNR improvement of +0.34dB over RegNeRF on the LLFF dataset and +0.53dB on the DTU dataset.
\end{enumerate}
\vspace{-15pt}

\section{Related Work}
\vspace{-5pt}
\subsection{Neural Radiance Fields}
\vspace{-5pt}
The introduction of Neural Radiance Fields (NeRF)\cite{mildenhall2020nerfrepresentingscenesneural} revolutionized 3D scene reconstruction by learning continuous volumetric representations through coordinate-based MLPs. 
Subsequent works have optimized NeRF across multiple dimensions: Instant-NGP\cite{mueller2022instant} and PlenOctrees\cite{yu2021plenoctreesrealtimerenderingneural} dramatically accelerate NeRF rendering via hybrid neural-voxel representations. NeRF\texttt{--}\cite{wang2021nerfmm} and UP-NeRF\cite{kim2023upnerf} eliminate the dependency on pre-calibrated camera poses by jointly optimizing pose estimation and radiance field reconstruction. Mip-NeRF\cite{barron2021mipnerfmultiscalerepresentationantialiasing} addresses aliasing artifacts by modeling conical frustums instead of rays, inherently preserving high-frequency details while unifying coarse and fine networks into a single multiscale MLP.

\vspace{-5pt}

\subsection{Sparse Views Novel View Synthesis}
\vspace{-5pt}
Recent approaches for sparse-view reconstruction fall into two categories:

\textbf{Pretraining-dependent approaches} leverage cross-scene priors learned from large datasets. MVSNeRF\cite{chen2021mvsnerffastgeneralizableradiance} integrates multi-view stereo geometry into NeRF through cost volume construction. PixelNeRF\cite{yu2021pixelnerfneuralradiancefields} encodes image features into a spatially aligned latent space for few-shot generalization. Stereo Radiance Fields (SRF)\cite{chibane2021stereoradiancefieldssrf} employs epipolar geometry constraints from stereo pairs to enhance reconstruction robustness. Despite their strong performance, these methods demand extensive multi-scene datasets and tend to generalize poorly to unseen categories or out-of-distribution domains.

\textbf{Regularization-driven optimization} methods impose scene-specific constraints during per-scene training: InfoNeRF\cite{kim2022infonerfrayentropyminimization} minimizes the entropy of each ray's density to enforce sparsity and ensures consistency across neighboring rays via a spatial smoothness constraint. DietNe\-RF\cite{jain2021puttingnerfdietsemantically} projects rendered patches and input views into CLIP's joint image-text space, enforcing semantic consistency via cosine similarity losses. RegNeRF\cite{niemeyer2021regnerfregularizingneuralradiance} introduces a dual-constraint framework which combines a geometry-aware TV loss on estimated depth maps, and an appearance-matching term that minimizes photometric warping errors between adjacent views. However, these global regularization strategies often over-smooth geometric boundaries, sacrificing fine structural details in pursuit of global consistency. In contrast, our EdgeNeRF introduces an edge-guided local regularization mechanism that explicitly preserves geometric discontinuities while enhancing consistency in non-edge regions.

\section{Method}
\subsection{Preliminaries}
\noindent\textbf{Neural Radiance Fields.} 
NeRF\cite{mildenhall2020nerfrepresentingscenesneural} represents a scene as a continuous function
\begin{equation}
{F}_{\Theta}:({x}, {d}) \rightarrow ({c}, \sigma),
\end{equation}
that maps 3D coordinates ${x}$ and viewing directions ${d} $ to color ${c}$ and volume density $\sigma$, $\Theta$ are the parameters of an 8-layer MLP with ReLU activations and residual connections.

% \vspace{0.5\baselineskip}

\noindent\textbf{Rendering.} For a camera ray ${r}(t) = {o} + t{d}$, where ${o}$ is the camera's optical center, ${d}$ is the unit direction vector from ${o}$ to the pixel, $t \in [t_{n}, t_{f}]$ denotes the distance along the ray, and $t_{n}, t_{f}$ are the near and far bounds of ray $r$. The pixel color is computed via volume rendering:
\begin{equation}
C({r})=\int_{t_n}^{t_f} T(t) \sigma({r}(t)) {c}({r}(t), {d}) \mathrm{d} t,
\label{formula:volume_rendering}
\end{equation}
where $T(t)=\exp \left(-\int_{t_n}^t \sigma({r}(s)) \mathrm{d} s\right)$ is the accumulated transmittance, $\sigma({r}(t))$ denotes volume density at point ${r}(t)$ and ${c}({r}(t), {d})$ denotes RGB color viewed from direction $d$ at point $r(t)$. The model is optimized by minimizing the photometric error between rendered and ground truth pixel colors:
\begin{equation}
    \mathcal{L}_{c} = \sum_{{r}\in\mathcal{R}}\|{C}(r) - C_\mathrm{GT}(r)\|_2^2
\label{formula:nerf_loss}
\end{equation}
where $C_\mathrm{GT}({r})$ denotes ground truth color and $\mathcal{R}$ is the set of all rays in one training epoch. 

\begin{figure}[hbt]
	\centering	
	\includegraphics[width=\linewidth]{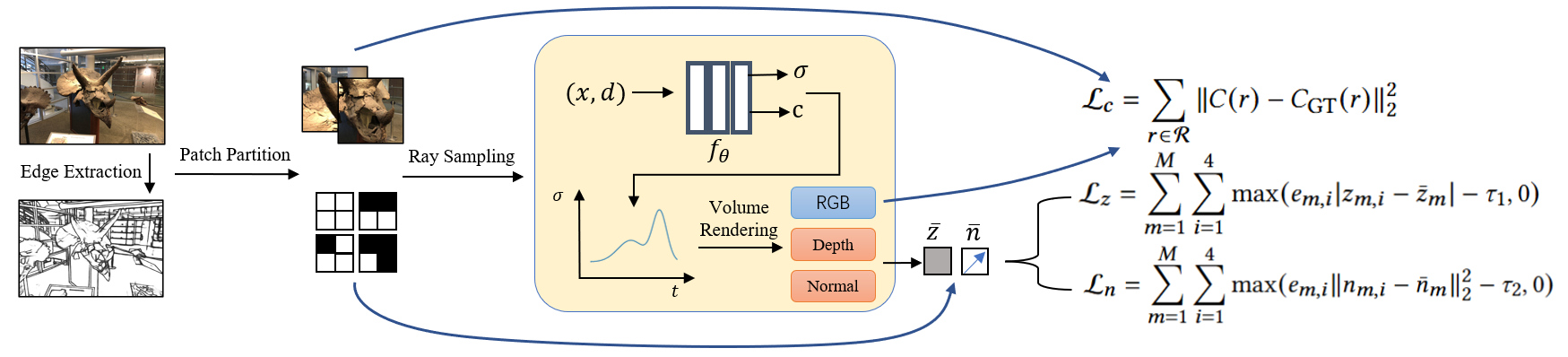}
	\caption{The pipeline of edge-guided sparse-view 3D reconstruction framework EdgeNeRF.}
    \label{fig:framework}
\vspace{-20pt}
\end{figure}

\subsection{Overview of EdgeNeRF}

The pipeline of edge-guided sparse-view 3D reconstruction framework EdgeNeRF is illustrated in Figure \ref{fig:framework}. EdgeNeRF operates on monocular RGB images with corresponding edge maps $\{({I}_i, {E}_i) ,i=1, \dots , N\}$. Unlike conventional NeRF's pixel-wise sampling, EdgeNeRF implements patch-based optimization using $2 \times 2$ patches to enhance spatial coherence. The framework derives depth $z$ and surface normals $n$ through differentiable rendering, regularized by edge-guided constraints to preserve geometric consistency and high-frequency details.

% The pipeline of edge-guided sparse-view 3D reconstruction framework EdgeNeRF is illustrated in Figure \ref{fig:framework}. EdgeNeRF operates on a sequence of monocular RGB images with corresponding edge maps $\{({I}_i, {E}_i) ,i=1, \dots , N\}$, where each RGB image ${I}_i \in \mathbb{R}^{H \times W \times 3}$ and its associated edge map ${E}_i \in \mathbb{R}^{H \times W}$ are captured from varying viewpoints, with $H$ and $W$ denoting the image height and width respectively. Departing from conventional NeRF's pixel-wise sampling approach, EdgeNeRF implements a patch-based optimization strategy that partitions input images into $2 \times 2$ patches and randomly samples a specified number of these patches during training to enhance spatial coherence. Through its differentiable rendering pipeline, the framework simultaneously derives both depth $d$ through ray termination probability integration and surface normals ${n}$ via spatial gradient computation, which are subsequently regularized using edge-guided constraints to maintain geometric consistency while preserving high-frequency details.

\subsection{Edge Extraction}

\begin{figure}[htbp]
\vspace{-10pt}
	\subfloat[Original Image]{\includegraphics[width=0.3\linewidth]{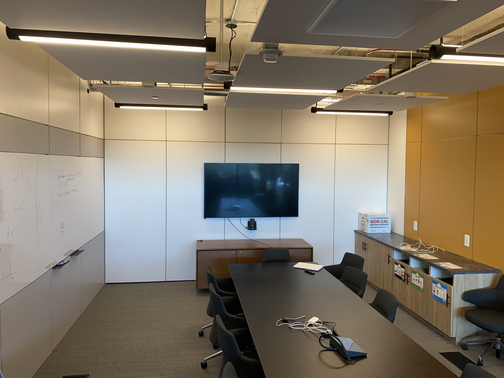}}%
	\hfill
	\subfloat[Edge Extraction by Canny]{\includegraphics[width=0.3\linewidth]{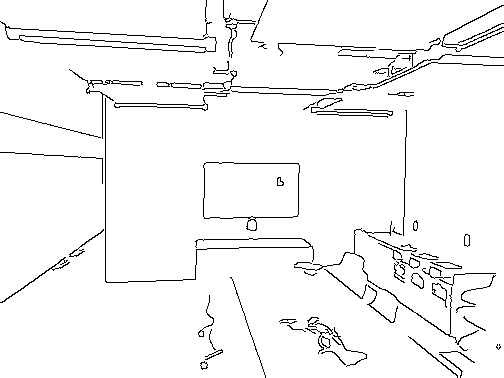}}%
	\hfill
	\subfloat[Edge Extraction by DexiNed]{\includegraphics[width=0.3\linewidth]{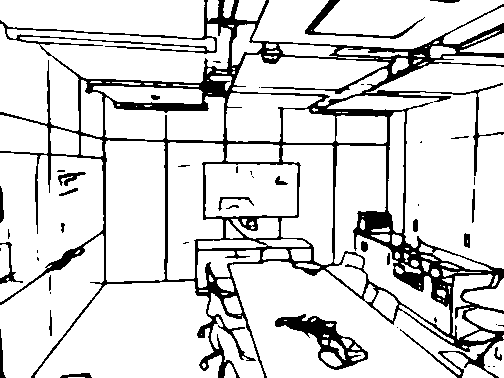}}
	\caption{Comparison on different edge extraction methods. (b) and (c) display binarized outputs from edge extraction. While the Canny operator fails to capture low-contrast edges (e.g., the back of the chair), DexiNed provides more complete and coherent edge maps, especially for fine structures.}
	\label{fig:edge_example}
\vspace{-15pt}
    
\end{figure}
Edges in images often arise from rapid intensity changes caused by object boundaries, material transitions, shadows, or surface texture variations. We adopt a simplified Lambertian model: suppose a scene surface $S$ with spatially varying albedo $\alpha({x})$, illuminated by a directional light source of intensity $I_0$ from direction ${l}$, the observed intensity at location ${x}$ is:
\begin{equation}
I({x}) = I_0 \alpha({x}) ({n}({x}) \cdot {l}),
\end{equation}
where $n(x)$ denotes the surface normal at location ${x}$.

An edge is typically identified when intensity variation exceeds a threshold: $\Delta I({x}) > \varepsilon$. From the Lambertian formulation, two key physical factors can lead to such intensity changes:

(\textbf{1}) A sharp change in surface orientation ($\Delta {n} > \varepsilon_n$), often caused by folds, corners, or creases;

(\textbf{2}) A change in material reflectance, i.e., a discontinuity in albedo ($\Delta \alpha({x}) > \varepsilon_\alpha$).

In addition, depth discontinuities (e.g., occlusion boundaries) often coincide with changes in visibility or surface orientation, contributing indirectly to edge formation.

In summary, abrupt changes in depth or surface normals serve as sufficient—though not necessary—conditions for edge formation. This justifies our regularization strategy: applying smoothness constraints only to non-edge regions preserves true geometric boundaries while improving reconstruction consistency.

% Image edges correspond to regions with significant local variations in pixel values, including discontinuities in color, brightness or texture, which commonly arise from underlying factors like object boundaries, material transitions, lighting changes or geometric discontinuities. We formally establish that depth discontinuities $(\frac{\partial d}{\partial x} \neq 0)$ and normal vector field variations $(\Delta {n})$ constitute sufficient but non-necessary conditions for edge formation through the following theoretical analysis: consider a continuous surface $S$ with albedo $\alpha$, where the observed image intensity $I({x})$ at pixel ${x}$ follows the Lambert's cosine law $I({x}) = I_0\alpha({x})({n}({x}) \cdot {l})$, where $\alpha({x})$ denotes the surface albedo, $I_0$ denotes incident light intensity, ${n}({x})$ denotes the surface normal and ${l}$ denotes the incident light direction. A necessary condition for edge formation is $\Delta I({x}) > \varepsilon$, which occurs when either $(1)$ the surface normal ${n}$ changes abruptly $(\Delta {n})$, or $(2)$ the depth $d$ exhibits discontinuity $(\frac{\partial d}{\partial x} \neq 0)$. However, edges may also emerge without geometric discontinuities when $\alpha({x})$ or $I_0$ changes sharply. 

In EdgeNeRF, edge extraction is a critical preprocessing step, where we derive corresponding edge maps $\{({E}_i) ,i=1, \dots , N\}$ from the input monocular RGB sequence $\{({I}_i) \allowbreak,i=1, \dots , N\}$. Traditional Canny edge detection\cite{canny1986computational} based on gradient thresholding, performs poorly under complex textures, motion blur, or low contrast. DexiNed\cite{dexined}, a deep learning-based edge detection method, yields more continuous and semantically meaningful edges compared to Canny's fragmented outputs, as illustrated in Figure \ref{fig:edge_example}.

After extracting the edge map, we binarize it as follows:
\begin{equation}
    {B}_i(x, y) = 
    \begin{cases}
    1, & if \ {E}_i(x, y) \geq \tau_{e}, \\
    0, & otherwise,
    \end{cases}
\end{equation}
where ${B}_i \in {\{0, 1\}}^{H \times W}$ is the binary edge map for the $i$-th image, ${E}_i(x, y)$ denotes the grayscale intensity at pixel position $(x,y)$ in the original edge image, and $\tau_e$ is the binarization threshold. 

To enhance edge continuity, we apply morphological dilation using a $3\times3$ all-ones kernel to expand the edge regions (where ${B}_i(x, y) = 0$), resulting in a refined edge map ${B}'_i$. During each training iteration, we sample $M$ $2\times2$ image patches $\{P_m^I\}_{m=1}^M$ along with their corresponding edge patches $\{P_m^{B'}\}_{m=1}^M$, with the key constraint that non-edge patches must belong to the same object surface to ensure spatial continuity of both depth and normal vectors, where $M$ represents the number of patches sampled per iteration.

\subsection{Edge-guided Depth Regularization}

\begin{figure}[htbp]
    \vspace{-5pt}
\centering
    \includegraphics[width=0.3\linewidth, frame]{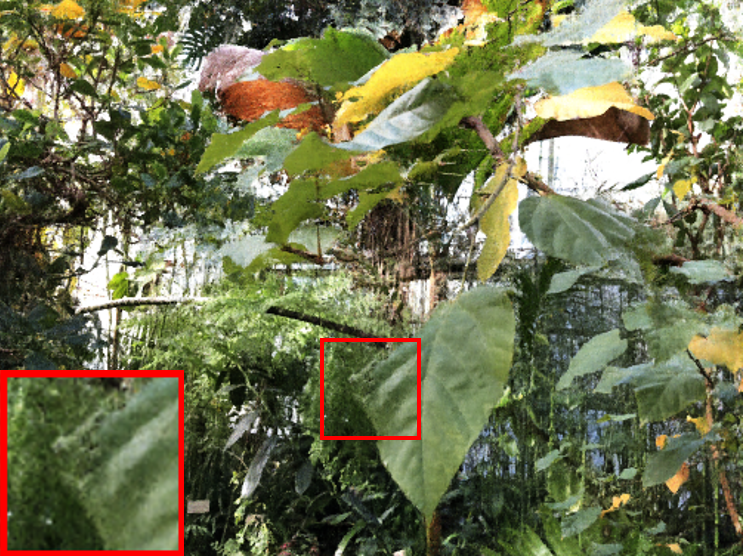}%
    \hspace{6pt}
    \includegraphics[width=0.3\linewidth, frame]{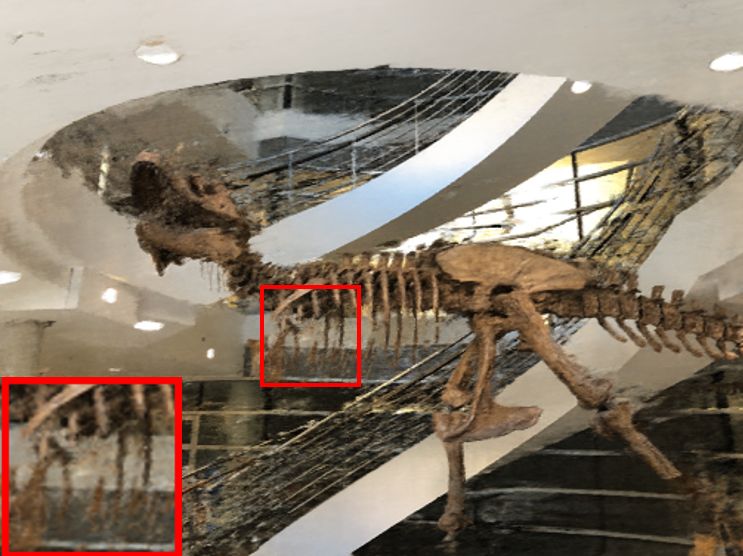}%
	\caption{Global depth smoothing of RegNeRF causes edge blurring and reconstruction failure.}
	\label{fig:regnerf-boarder-fail}
    \vspace{-25pt}
\end{figure}

While conventional global depth regularization methods (e.g., RegNeRF\cite{niemeyer2021regnerfregularizingneuralradiance}) enhance geometric quality and improve novel view synthesis, such coarse-grained smoothing strategies adversely affect the optimization process. The underlying issue is that real-world scenes exhibit local smoothness in geometry but also contain significant depth discontinuities at object boundaries and structural variations. A simple global smoothing approach fails to accurately capture these geometric characteristics, and instead weakens the model's ability to preserve fine details. As illustrated in Figure \ref{fig:regnerf-boarder-fail}, RegNeRF introduces noticeable geometric blurring in depth discontinuity regions (typically corresponding to object boundaries). 

Motivated by the finding that blurring phenomenon at geometric boundaries fundamentally stems from neglecting local structural information, we propose an edge-guided depth regularization framework that enforces smooth depth transitions within non-edge regions.

Analogous to the pixel color computation via volume rendering in Eq. \eqref{formula:volume_rendering}, the depth of pixel corresponding to ray $r$ is calculated as:
\begin{equation}
    z({r})=\int_{t_n}^{t_f} T(t) \sigma({r}(t)) t \mathrm{d} t.
\end{equation}

For the $m$-th image patch ${P}^{{I}}_m$, its weighted average depth is computed by:
\begin{equation}
    \bar{z}_m =\frac{\sum_{i=1}^{4}e_{m,i} z_{m,i}}{\sum_{i=1}^{4} e_{m,i}},
\end{equation}
where $e_{m,i} \in \{0, 1\}$ denotes the binary edge indicator from edge patch ${P}^{{B}^{'}}_m$ and $z_{m,i}$ denotes the depth value of the $i$-th pixel in ${P}^{{I}}_m$. The depth regularization loss implements selective smoothing over non-edge regions:
\begin{equation}
    \mathcal{L}_z =\sum_{m=1}^{M} \sum_{i=1}^{4} \max (e_{m,i}|z_{m,i} - \bar{z}_m| - \tau_1, 0),
\end{equation}
where $\tau_1$ is a preset depth smoothing tolerance threshold. In edge regions ($e_{m,i} = 0$), the corresponding depth values are excluded from both the weighted average and loss computation, preserving edge details. In contrast, in non-edge regions ($e_{m,i} = 1$), the depth values are included, enabling local smoothing and enhancing reconstruction quality.

\vspace{-5pt}
\vspace{-5pt}

\subsection{Edge-guided Normal Regularization}
Although depth regularization effectively enforces geometric continuity, a complete 3D scene representation requires precise estimation of the normal vector field. Since surface normals encode fundamental geometric orientation, their spatial consistency becomes a critical determinant of reconstruction quality: in non-edge regions (continuous surfaces), adjacent normals should exhibit smooth transitions, whereas at edge regions (depth discontinuities), abrupt changes in normal direction constitute the intrinsic characteristic of geometric boundaries. To properly constrain this spatially adaptive normal field, we need to develop a regularization approach that simultaneously preserves local smoothness and maintains geometric edges.

Current NeRF implementations derive surface normal via either $(1)$ direct unit vector prediction using MLPs at 3D coordinates\cite{bi2020neural,zhang2021nerfactor}, or $(2)$ computation through the gradient of volume density with respect to 3D positions\cite{boss2021nerd,verbin2022ref}. We adopt the latter approach, defining the normal vector field through volume density gradients:
\begin{equation}
    {n}({r}(t))=-\frac{\nabla \sigma({r}(t))}{\|\nabla \sigma({r}(t))\|}.
\end{equation}
Following the volume rendering paradigm, the expected normal is calculated as:
\begin{equation}
    {n}({r})=\int_{t_n}^{t_f} T(t) \sigma({r}(t)) {n}({r}(t)) \mathrm{d} t.
\end{equation}
For the $m$-th image patch ${P}^{{I}}_m$, the weighted average normal is:
\begin{equation}
    \bar{{n}}_m =\frac{\sum_{i=1}^{4}e_{m,i} {n}_{m,i}}{\sum_{i=1}^{4} e_{m,i}},
\end{equation}
where ${n}_{m,i}$ is the normal of the $i$-th pixel in ${P}^{{I}}_m$.
The normal regularization loss for non-edge regions enforces piecewise smoothness:
\begin{equation}
    \mathcal{L}_n =\sum_{m=1}^{M} \sum_{i=1}^{4} \max (e_{m,i}\|{n}_{m,i} - \bar{{n}}_m\|_2^2 - \tau_2, 0),
\end{equation}
where $\tau_2$ is a preset normal variation tolerance threshold.

\subsection{Optimization}
% Given a sequence of RGB images and pre-processed edge images $\{({I}_i, {B}'_i) ,i=1, \dots , N\}$, along with a multilayer perceptron ${F}_{{\Theta}}$, camera intrinsic matrix $K$ and extrinsic matrix $T$ corresponding to the RGB capture, our method optimizes model parameters $\Theta$ through a joint loss function that simultaneously fits observed data and satisfies geometric prior constraints:

Our method optimizes NeRF model using a joint loss function that simultaneously fits observed data and enforces geometric prior constraints:
\begin{equation}
    \mathcal{L}=\lambda_1 \mathcal{L}_c + \lambda_2 \mathcal{L}_z + \lambda_3 \mathcal{L}_n,
    \label{formula:edgenerf_total_loss}
\end{equation}
where $\lambda_j(j=1,2,3)$ denotes composite weight coefficients and $\mathcal{L}_c$ denotes photometric loss as illustrated in Eq. \eqref{formula:nerf_loss}.

During each training iteration, we randomly select one image from the $N$ training samples and extract $n$ patches of size $2\times2$ for optimization. Following Mip-NeRF's projection cone sampling strategy, we cast light cones per pixel using camera intrinsics and extrinsics, sample within these frustums, and apply integrated positional encoding to enhance spatial representation. For each ray, we compute volume density and color at sampled points. The network parameters are updated by backpropagating the total loss in Eq. \eqref{formula:edgenerf_total_loss}. 
This optimization framework enables learning a geometrically consistent radiance field that significantly improves novel view synthesis quality under sparse-view conditions.

\section{Experiments}

\begin{table}[htbp]
\vspace{-15pt}

    \caption{Quantitative comparison on the LLFF dataset. Our EdgeNeRF achieves the best results in all metrics under three input views. We reproduce RegNeRF without the appearance regularization module mark as \dag RegNeRF. The best, second-best, and third-best entries are marked in red, orange, and yellow, respectively.}
    \label{table:edgenerf_llff_total}
    
    \centering
    \begin{tabular}{l|>{\centering \arraybackslash}p{4cm}|c|c|c}
    \toprule
                 & Setting                                  & PSNR$\uparrow$  & SSIM$\uparrow$  & LPIPS$\downarrow$  \\ 
    \midrule
    SRF          & \multirow{3}{4cm}{\centering Training on DTU}            & 12.34 & 0.250 & 0.591  \\
    pixelNeRF    &                                          & 7.93  & 0.272 & 0.682  \\
    MVSNeRF      &                                          & 17.25 & 0.557 & 0.356  \\ 
    \midrule
    SRF ft       & \multirow{3}{4cm}{\centering Training on DTU and Optimized per Scene} & 17.07 & 0.436 & 0.529  \\
    pixelNeRF ft &                                          & 16.17 & 0.438 & 0.512  \\
    MVSNeRF ft   &                                          & 17.88 & 0.584 & 0.327  \\ 
    \midrule
    Mip-NeRF     & \multirow{5}{4cm}{\centering Optimized per Scene}               & 14.62 & 0.351 & 0.495  \\
    DietNeRF     &                                          & 14.94 & 0.370 & 0.496  \\
    RegNeRF      &                                          & \cellcolor{myorange}{19.08} & \cellcolor{myyellow}{0.587} & \cellcolor{myorange}{0.336}  \\
    \dag RegNeRF      &                             & \cellcolor{myyellow}{18.83} & \cellcolor{myorange}{0.673} & \cellcolor{myyellow}{0.346}  \\
    \textbf{EdgeNeRF} &                          & \cellcolor{myred}{19.42} & \cellcolor{myred}{0.699} & \cellcolor{myred}{0.317}  \\
    \bottomrule
    \end{tabular}
    
\vspace{-5pt}
\end{table}

\noindent\textbf{Datasets \& Metrics.} We evaluate our method on two established benchmarks: LLFF\cite{llff} and DTU\cite{dtu}. The LLFF dataset comprises 8 real-world forward-facing scenes, where each scene contains between 20 to 62 images. Camera parameters are estimated via COLMAP\cite{colmap}. Adopting the experimental protocol from RegNeRF, we reserve every eighth image for testing while uniformly selecting sparse training views from the remaining images. All images are downsampled to $504\times378$ resolution during training. The DTU dataset contains 124 object-centric scenes under seven controlled lighting conditions. Following PixelNeRF\cite{yu2021pixelnerfneuralradiancefields}, we employ their standard 15-scene subset for evaluation. Notably, DTU scenes feature simplified backgrounds (either white tabletops or black backdrops), to ensure unbiased evaluation, we implement background masking during quantitative assessment as specified in RegNeRF's evaluation protocol. We adopt PSNR, SSIM\cite{1284395} and LPIPS\cite{zhang2018unreasonableeffectivenessdeepfeatures} as the evaluation metrics.

\vspace{0.3\baselineskip}

\noindent\textbf{Implementation Details.} 
We implement EdgeNeRF based on RegN\-eRF\cite{niemeyer2021regnerfregularizingneuralradiance}, with all methods developed using the JAX framework\cite{bradbury2018jax}. For optimization, we follow RegNeRF's hyperparameter configuration, while the batch size is set to 4096 (corresponding to $M=1024$ sampled patches). 

% Primary experiments are conducted on an 80GB GPU-A800, with additional experiments validated on 24GB RTX-3090Ti and 12GB GTX-1080Ti GPUs, where we reduce the batch size to 1024 due to memory limitations. We follow the same training schedule as RegNeRF, using 69768 iterations per scene.

The parameter configuration remains consistent across experiments: $\tau_e = 125$, $\lambda_1=1$, $\lambda_2=0.1$, $\tau_1=10^{-4}$, $\tau_2=0$. Dataset-specific adjustments include $\lambda_3=0.1$ for LLFF and $\lambda_3=0.001$ for DTU evaluations.

\vspace{-8pt}

\begin{figure}[htbp]
    \centering
    \setlength{\tabcolsep}{4pt} % 紧凑列间距
    \renewcommand{\arraystretch}{1.1}
    
    % 定义统一高度（保持原始比例的关键）
    \newlength{\imgheight}
    \setlength{\imgheight}{2.5cm} % 可根据需要调整
    
    % 第一行：带标题
    \begin{tabular}{@{}c@{\hspace{6pt}}c@{\hspace{6pt}}c@{}}
        \scriptsize Ground Truth & \scriptsize RegNeRF & \scriptsize EdgeNeRF \\
        \includegraphics[height=\imgheight,width=0.18\linewidth,keepaspectratio]{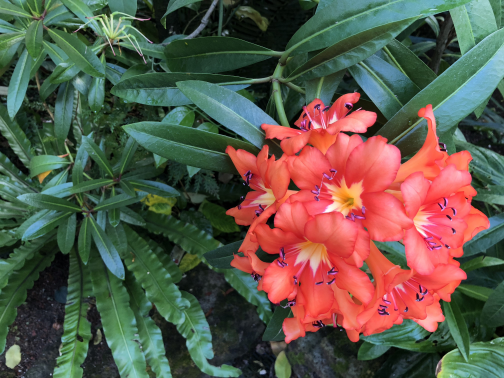} &
        \includegraphics[height=\imgheight,width=0.36\linewidth,keepaspectratio]{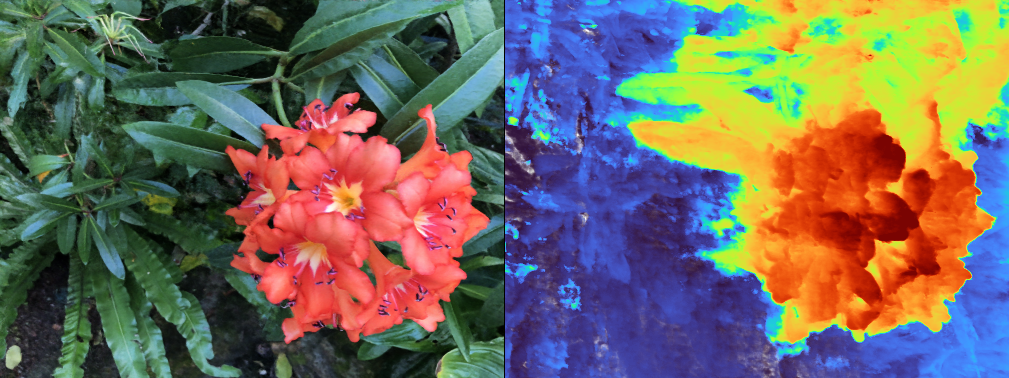} &
        \includegraphics[height=\imgheight,width=0.36\linewidth,keepaspectratio]{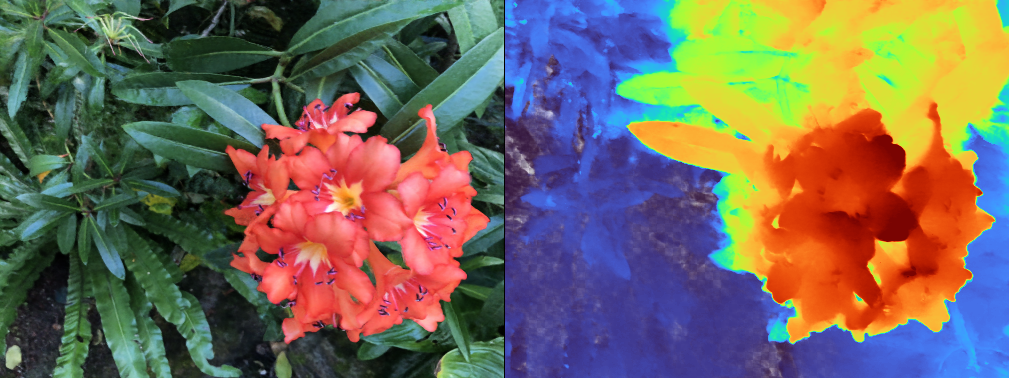}
    \end{tabular}
    
    \vspace{4pt}
    
    % 第二行
    \begin{tabular}{@{}c@{\hspace{6pt}}c@{\hspace{6pt}}c@{}}
        \includegraphics[height=\imgheight,width=0.18\linewidth,keepaspectratio]{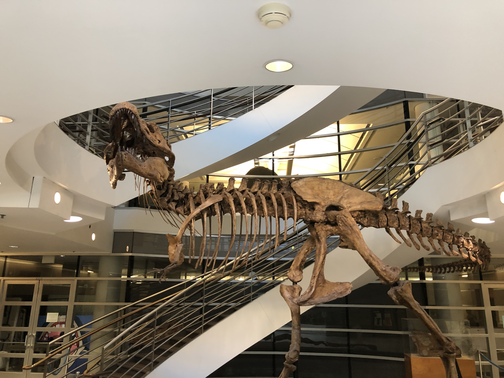} &
        \includegraphics[height=\imgheight,width=0.36\linewidth,keepaspectratio]{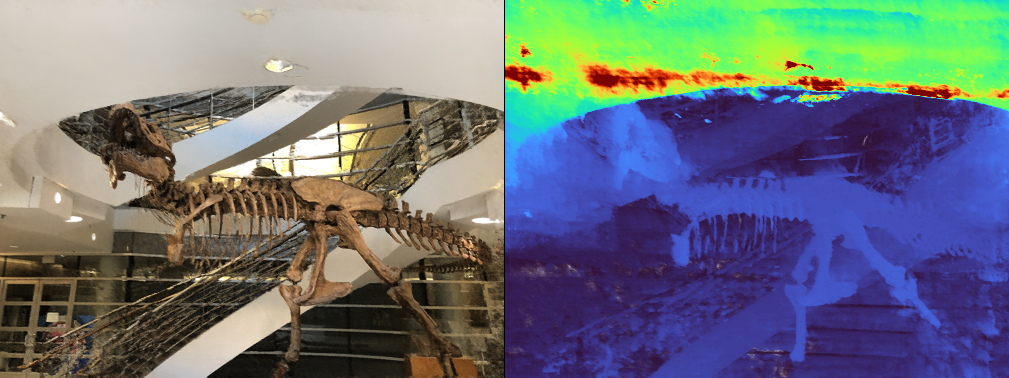} &
        \includegraphics[height=\imgheight,width=0.36\linewidth,keepaspectratio]{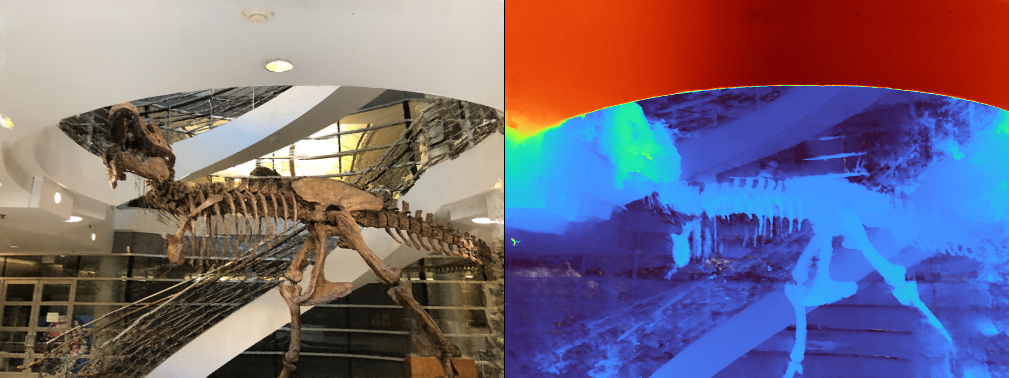}
    \end{tabular}
    
    % \vspace{4pt}
    
    % % 第三行
    % \begin{tabular}{@{}c@{\hspace{6pt}}c@{\hspace{6pt}}c@{}}
    %     \includegraphics[height=\imgheight,width=0.18\linewidth,keepaspectratio]{Fig/room_004.png} &
    %     \includegraphics[height=\imgheight,width=0.36\linewidth,keepaspectratio]{Fig/regnerf_room_004.png} &
    %     \includegraphics[height=\imgheight,width=0.36\linewidth,keepaspectratio]{Fig/edgenerf_room_004.png}
    % \end{tabular}
    
    \caption{Visual comparisons on the LLFF dataset with three input views. EdgeNeRF demonstrates superior performance in handling complex lighting conditions (flower scene) and fine geometric details (T-Rex model), with +0.41dB PSNR improvement over RegNeRF.}
    \label{fig:llff_qualitative_result}
\vspace{-15pt}
    
\end{figure}

\subsection{Comparison on LLFF}

Table \ref{table:edgenerf_llff_total} presents quantitative comparisons on the LLFF dataset using three input views. Results for MVSNeRF, PixelNeRF, and SRF are from \cite{niemeyer2021regnerfregularizingneuralradiance}. These methods are pre-trained on the DTU dataset due to the limited number of LLFF scenes, then undergo additional per-scene optimization ("ft") during testing. We also reproduce RegNeRF without the appearance regularization module as our comparison backbone, as the official RegNeRF implementation lacks this module.

Results demonstrate EdgeNeRF's superiority over most existing approaches. Figure \ref{fig:llff_qualitative_result} provides qualitative comparisons between EdgeNeRF and RegNeRF. EdgeNeRF significantly reduces floating artifacts and produces smoother depth estimates, whereas RegNeRF's depth maps exhibit noticeable inaccuracies, demonstrating the efficacy of our method.

\begin{table}[htbp]
\vspace{-18pt}

    \caption{Quantitative comparison on the DTU dataset with three input views. \dag RegNeRF: w/o. appearance regularization. The best, second-best, and third-best entries are marked in red, orange, and yellow, respectively.}
    \label{table:edgenerf_dtu_total}
    
    \centering
    \begin{tabular}{l|>{\centering \arraybackslash}p{4cm}|c|c|c} 
    \toprule
                 & Setting                          & PSNR$\uparrow$  & SSIM$\uparrow$  & LPIPS$\downarrow$  \\ 
    \midrule
    SRF          & \multirow{3}{4cm}{\centering Training on DTU}    & 15.32 & 0.671 & 0.304  \\
    pixelNeRF    &                                  & 16.82 & 0.695 & 0.270  \\
    MVSNeRF      &                                  & 18.63 & \cellcolor{myyellow}{0.769} & \cellcolor{myyellow}{0.197}  \\ 
    \midrule
    SRF ft       & \multirow{3}{4cm}{\centering Training on DTU and Optimized per Scene} & 15.68 & 0.698 & 0.281  \\
    pixelNeRF ft &                                  & \cellcolor{myorange}{18.95} & 0.710 & 0.269  \\
    MVSNeRF ft   &                                  & 18.54 & \cellcolor{myyellow}{0.769} & \cellcolor{myyellow}{0.197}  \\ 
    \midrule
    Mip-NeRF     & \multirow{5}{4cm}{\centering Optimized per Scene}       & 8.68  & 0.571 & 0.353  \\
    Diet-NeRF    &                                  & 11.85 & 0.633 & 0.314  \\
    RegNeRF      &                                  & \cellcolor{myyellow}{18.89} & 0.745 & \cellcolor{myred}{0.190}  \\
    \dag RegNeRF      &                     & 18.55 & \cellcolor{myorange}{0.811} & \cellcolor{myorange}{0.193}  \\
    \textbf{EdgeNeRF} &                             & \cellcolor{myred}{19.42} & \cellcolor{myred}{0.828} & 0.205  \\
    \bottomrule
    \end{tabular}
\vspace{-20pt}
\end{table}

\subsection{Comparison on DTU}
\begin{figure}[htbp]
\vspace{-5pt}
    \centering
    \setlength{\tabcolsep}{4pt} % 减少列间距
    \renewcommand{\arraystretch}{1.2} % 增加行高
    
    % 第一行：带标题的图片
    \begin{tabular}{@{}c@{\hspace{6pt}}c@{\hspace{6pt}}c@{}}
        \scriptsize Ground Truth & \scriptsize RegNeRF & \scriptsize EdgeNeRF \\
        \includegraphics[width=0.18\linewidth,height=2.2cm,keepaspectratio]{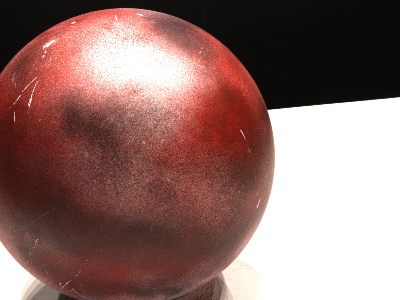} &
        \includegraphics[width=0.36\linewidth,height=2.2cm,keepaspectratio]{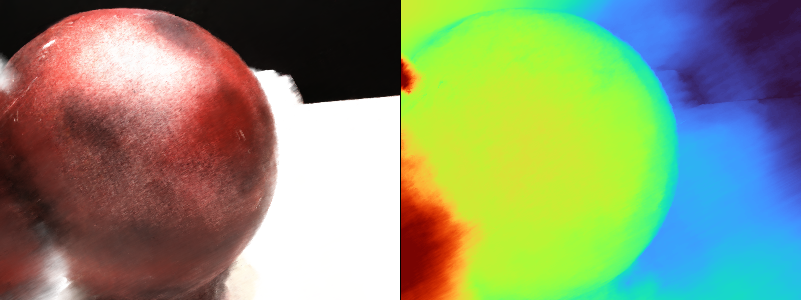} &
        \includegraphics[width=0.36\linewidth,height=2.2cm,keepaspectratio]{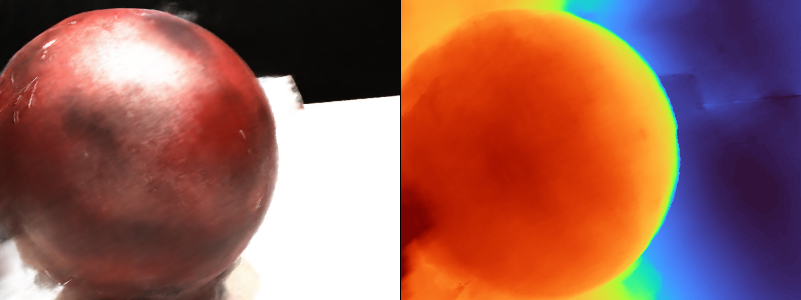}
    \end{tabular}
    
    \vspace{4pt}
    
    % 第二行：纯图片
    \begin{tabular}{@{}c@{\hspace{6pt}}c@{\hspace{6pt}}c@{}}
        \includegraphics[width=0.18\linewidth,height=2.2cm,keepaspectratio]{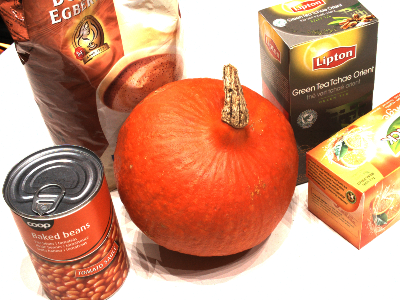} &
        \includegraphics[width=0.36\linewidth,height=2.2cm,keepaspectratio]{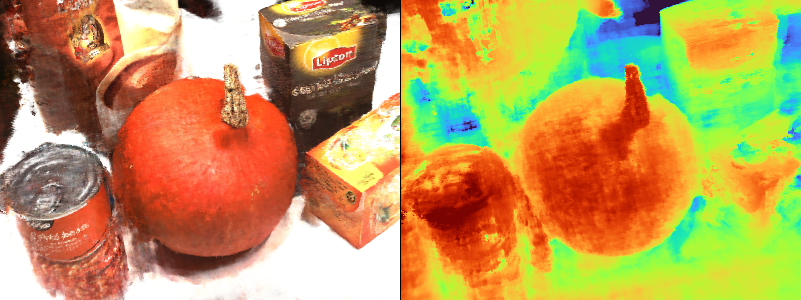} &
        \includegraphics[width=0.36\linewidth,height=2.2cm,keepaspectratio]{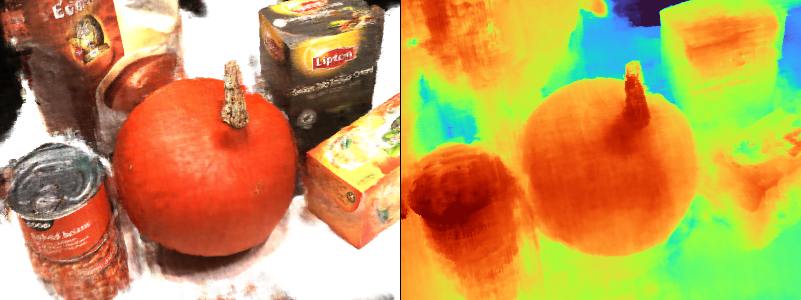}
    \end{tabular}
    
    % \vspace{4pt}
    
    % % 第三行：纯图片
    % \begin{tabular}{@{}c@{\hspace{6pt}}c@{\hspace{6pt}}c@{}}
    %     \includegraphics[width=0.18\linewidth,height=2.2cm,keepaspectratio]{Fig/scan40.png} &
    %     \includegraphics[width=0.36\linewidth,height=2.2cm,keepaspectratio]{Fig/regnerf_scan40.png} &
    %     \includegraphics[width=0.36\linewidth,height=2.2cm,keepaspectratio]{Fig/edgenerf_scan40.png}
    % \end{tabular}
    
    \caption{Visual comparisons on the DTU dataset with three input views. EdgeNeRF demonstrates clear advantages in geometric reconstruction, significantly reducing artifacts and producing lower-noise depth estimates.}
    \label{fig:dtu_qualitative_result}
\vspace{-15pt}
\end{figure}

Table \ref{table:edgenerf_dtu_total} presents the quantitative comparison on DTU dataset using three input views, with experimental configurations consistent with those on LLFF dataset. Following RegNeRF's protocol, we apply object masks to render images during evaluation to prevent performance penalization from erroneous background predictions. Figure \ref{fig:dtu_qualitative_result} shows qualitative comparisons between EdgeNeRF and RegNeRF. Although primarily optimized for multi-object scenes, EdgeNeRF maintains competitive performance in single-object reconstruction, significantly enhancing geometric accuracy while reducing artifacts

% Table \ref{table:edgenerf_dtu_total} presents the quantitative comparison on DTU dataset using three input views, with experimental configurations consistent with those on LLFF dataset. Following RegNeRF's protocol, we apply object masks to render images during evaluation to prevent performance penalization from erroneous background predictions. The results demonstrate our method's superior performance, EdgeNeRF achieves \textbf{+0.87dB} PSNR and \textbf{+0.083} SSIM improvement over RegNeRF without appearance regularization; \textbf{+0.53dB} PSNR and \textbf{+0.017} SSIM improvement over the full RegNeRF.

% Figure \ref{fig:dtu_qualitative_result} shows qualitative comparisons between EdgeNeRF and RegNeRF. Although primarily optimized for multi-object scenes, EdgeNeRF maintains competitive performance in single-object reconstruction, significantly enhancing geometric accuracy while reducing artifacts.

% \subsection{Computational Overhead}
% We conduct 100 training iterations on LLFF with a batch size of 4096 using an A800 GPU to evaluate the computational overhead of our method, with the timing results summarized in Table \ref{table:edgenerf_train_time}. To ensure accurate measurement of actual training time, we excluded the initial JAX compilation phase from our timing calculations. Our experiments reveal that while depth regularization introduces negligible additional computational cost, normal regularization increases training time by approximately $0.3\times$ due to the computationally intensive gradient operations required for normal estimation. This analysis demonstrates that our depth regularization approach can be readily integrated into existing frameworks without substantially impacting training efficiency.

% \begin{table}[htbp]
% \vspace{-15pt}
%     \caption{Comparative training time ratio of the regularization methods. \dag RegNeRF: w/o. appearance regularization. All training times are normalized relative to \dag RegNeRF(1.0$\times$)}
%     \label{table:edgenerf_train_time}
%     \centering
%     \begin{tabular}{c|c}
%         \toprule
%             & Training time ratio \\
%         \midrule
%         \dag RegNeRF & 1.0$\times$ \\
%         w/ depth regularization & 1.003$\times$ \\
%         w/ normal regularization & 1.338$\times$ \\
%         \bottomrule
%     \end{tabular}
% \vspace{-15pt}
% \end{table}

\subsection{Ablation Studies and Further Analysis}

\noindent\textbf{Effectiveness of Depth and Normal Regularization.}
To validate the effectiveness of individual components in our method, we conduct ablation studies on the LLFF dataset. Quantitative results are presented in Table \ref{table:edgenerf_ablation}. It is observed that without the depth or normal regularization, EdgeNeRF degrades on PSNR, SSIM and LPIPS, which demonstrates the effectiveness of depth and normal regularization.

\begin{table}[htbp]
\vspace{-20pt}
    \caption{Ablation studies of EdgeNeRF on the LLFF dataset. \dag RegNeRF: w/o. appearance regularization.}
    \label{table:edgenerf_ablation}
    \centering
    \begin{tabular}{c|ccc}
    \toprule
    & PSNR$\uparrow$ & SSIM$\uparrow$ & LPIPS$\downarrow$ \\
    \midrule
    \dag RegNeRF & 18.83 & 0.673 & 0.346 \\
    \midrule
    w/o. depth regularization & 19.15 & 0.683 &  0.343 \\
    w/o. normal regularization & 19.30 & 0.695 & 0.318 \\
    EdgeNeRF & \textbf{19.42} & \textbf{0.699} & \textbf{0.317} \\
    \bottomrule
\end{tabular}
\vspace{-15pt}
\end{table}

% \vspace{0.3\baselineskip}
\vspace{-22pt}
\begin{table}[htbp]
    \caption{Comparison on different edge extraction methods. \dag RegNeRF: w/o. appearance regularization.}
    \label{table:edgenerf_edge_extract_contrast}
    \centering
    \begin{tabular}{c|ccc}
    \toprule
    & PSNR$\uparrow$ & SSIM$\uparrow$ & LPIPS$\downarrow$ \\
    \midrule
    \dag RegNeRF & 18.83 & 0.673 & 0.346 \\
    \midrule
    Canny & 19.40 & 0.697 &  0.335 \\
    Dexined & \textbf{19.42} & \textbf{0.699} & \textbf{0.317} \\
    \bottomrule
\end{tabular}
\vspace{-10pt}
\end{table}

% \noindent\textbf{Effectiveness of Depth and Normal Regularization.}
% We first assess the performance using only normal regularization. Results demonstrate that using only edge-guided normal regularization achieves a 0.32dB improvement in PSNR, 0.01 gain in SSIM and 0.003 reduction in LPIPS. Visual analysis confirms significant improvements in normal map quality: the map exhibits better smoothness while preserving sharper boundaries. And the geometric estimation (depth map) already surpasses RegNeRF even without depth regularization.

% We then evaluate the performance using depth regularization alone. Results show that our edge-guided depth regularization delivers more substantial improvements: +0.47dB PSNR, +0.022 SSIM and -0.028 LPIPS. Visual inspection reveals that depth regularization not only significantly enhances geometric accuracy but also indirectly improves normal estimation, confirming the intrinsic correlation between depth and normal estimation quality.

\noindent\textbf{Edge Extraction Methods.} In Table \ref{table:edgenerf_edge_extract_contrast}, we further compare different edge extraction methods (Canny $vs.$ DexiNed) using three input views on the LLFF dataset. Results indicate that both edge extractors outperform the backbone framework, even with the lightweight Canny operator, our method maintains competitive performance without significant degradation. This finding further demonstrates the flexibility and efficiency of our work.

\noindent\textbf{Parameters Study.} 
We conduct a sensitivity analysis on EdgeNeRF's hyperparameters $\lambda_2$ (depth regularization weight coefficient) and $\lambda_3$ (normal regularization weight coefficient) on the LLFF dataset. Five experimental configurations are designed to evaluate these parameters, with results documented in Table \ref{table:edgenerf_llff_lambda2_sensitive} and Table \ref{table:edgenerf_llff_lambda3_sensitive}. Both hyperparameters exhibit similar behavior: model performance initially improves then declines as weight coefficients increase, with both achieving optimal performance at a value of 0.1. The results demonstrate that both regularization terms require conservative initialization (start with $\lambda<0.01$), and progressive optimization strategies should be employed to prevent regularization from overriding primary reconstruction objectives.

\vspace{-20pt}

\begin{table}
    \centering
    \caption{The impact of $\lambda_2$ on the LLFF dataset}
    \label{table:edgenerf_llff_lambda2_sensitive}
    \begin{tabular}{l|ccccc}
    \toprule
                    & 0.01  & 0.05  & 0.1            & 0.5   & 1      \\
    \midrule
    PSNR$\uparrow$  & 19.11 & 19.16 & \textbf{19.42} & 19.40 & 19.35  \\
    SSIM$\uparrow$  & 0.680 & 0.684 & \textbf{0.699} & 0.693 & 0.684  \\
    LPIPS$\downarrow$ & 0.357 & 0.347 & \textbf{0.317} & 0.332 & 0.342 \\
    \bottomrule
\end{tabular}
\end{table}

\vspace{-37pt}

\begin{table}
    \centering
    \caption{The impact of $\lambda_3$ on the LLFF dataset}
    \label{table:edgenerf_llff_lambda3_sensitive}
    \begin{tabular}{l|ccccc}
    \toprule
                    & 0.01  & 0.05  & 0.1   & 0.5   & 1      \\
    \midrule
    PSNR$\uparrow$  & 19.21 & 19.19 & \textbf{19.42} & 19.18 & 19.23  \\
    SSIM$\uparrow$  & 0.686 & 0.687 & \textbf{0.699} & 0.688 & 0.685  \\
    LPIPS$\downarrow$ & 0.339 & 0.340 & \textbf{0.317} & 0.335 & 0.338 \\
    \bottomrule
\end{tabular}
\end{table}

\vspace{-10pt}

% \begin{table}
%     \centering
%     \caption{The impact of $\lambda_2$ on the DTU dataset}
%     \label{table:edgenerf_dtu_lambda2_sensitive}
%     \begin{tabular}{l|ccccc} 
%     \toprule
%           & 0.01  & 0.05  & 0.1   & 0.5   & 1.0    \\ 
%     \midrule
%     PSNR$\uparrow$  & 19.07 & 19.06 & \textbf{19.42} & 18.19 & 16.25  \\
%     SSIM$\uparrow$  & 0.820 & 0.821 & \textbf{0.828} & 0.816 & 0.790  \\
%     LPIPS$\downarrow$ & 0.210 & 0.209 & \textbf{0.205} & 0.223 & 0.249  \\
%     \bottomrule
%     \end{tabular}
% \end{table}

% \begin{table}
%     \centering
%     \caption{The impact of $\lambda_3$ on the DTU dataset}
%     \label{table:edgenerf_dtu_lambda3_sensitive}
%     \begin{tabular}{l|ccccc} 
%     \toprule
%                                 & 0.001 & 0.005 & 0.01  & 0.05  & 0.1    \\ 
%     \midrule
%     PSNR$\uparrow$              & 19.17 & 19.21 & \textbf{19.42} & 19.08 & 18.99  \\
%     SSIM$\uparrow$              & 0.825 & 0.823 & \textbf{0.828} & 0.820 & 0.820  \\
%     LPIPS$\downarrow$           & 0.209 & 0.210 & \textbf{0.205} & 0.212 & 0.213  \\
%     \bottomrule
%     \end{tabular}
% \end{table}

\noindent\textbf{Computational Overhead.}
We evaluate our method's computational overhead in Table \ref{table:edgenerf_train_time}. To ensure accurate measurement of actual training time, we excluded the initial JAX compilation phase from our timing calculations. Our experiment reveals that depth regularization introduces negligible cost, demonstrating that our depth regularization approach can be readily integrated into existing frameworks without substantially impacting training efficiency.
% We conduct 100 training iterations on LLFF with a batch size of 4096 using an A800 GPU to evaluate the computational overhead of our method, with the timing results summarized in Table \ref{table:edgenerf_train_time}. To ensure accurate measurement of actual training time, we excluded the initial JAX compilation phase from our timing calculations. Our experiments reveal that while depth regularization introduces negligible additional computational cost, normal regularization increases training time by approximately $0.3\times$ due to the computationally intensive gradient operations required for normal estimation. This analysis demonstrates that our depth regularization approach can be readily integrated into existing frameworks without substantially impacting training efficiency.

\vspace{-20pt}

\begin{table}[htbp]
    \caption{Comparative training time ratio of the regularization methods. \dag RegNeRF: w/o. appearance regularization. All training times are normalized relative to \dag RegNeRF(1.0$\times$)}
    \label{table:edgenerf_train_time}
    \centering
    \begin{tabular}{c|c}
        \toprule
            & Training time ratio \\
        \midrule
        \dag RegNeRF & 1.0$\times$ \\
        w/ depth regularization & 1.003$\times$ \\
        w/ normal regularization & 1.338$\times$ \\
        \bottomrule
    \end{tabular}
\end{table}

\vspace{-10pt}

% \subsection{Integration into Other Methods}
\noindent\textbf{Integration into SparseNeRF.} 
To show our edge-guided depth regularization is effective and easy to integrate, we replaced SparseNeRF's \cite{sparsenerf} spatial continuity module. We upgraded its depth consistency component to Depth Anything V2 \cite{yang2024depth} for better accuracy and simplified deployment with a global depth ordering strategy.

\begin{figure}[htbp]
\centering
    \subfloat[SparseNeRF]{\includegraphics[width=0.3\linewidth, frame]{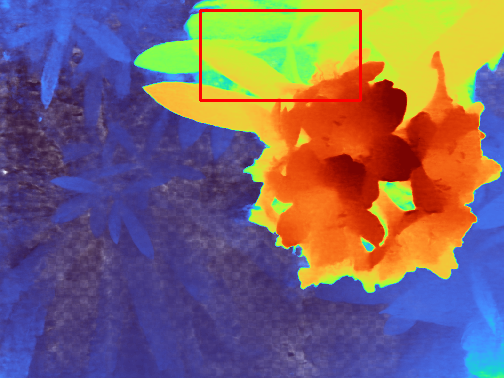}}%
    \hspace{6pt}
    \subfloat[SparseNeRF + our depth regularization]{\includegraphics[width=0.3\linewidth, frame]{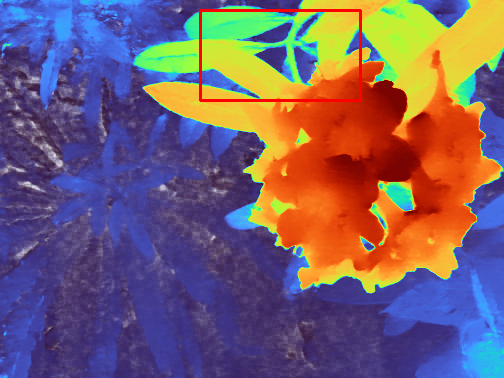}}%
	\caption{Qualitative comparison of SparseNeRF with depth regularization.}
	\label{fig:edgenerf-rank_vs_sparsenerf}
\end{figure}

For fair comparison, we retrained SparseNeRF under identical conditions. Evaluations (Table \ref{table:edgenerf_vs_sparsenerf} and Figure \ref{fig:edgenerf-rank_vs_sparsenerf}) reveal that combining our method with SparseNeRF significantly improves geometric estimation. Our approach complements SparseNeRF's global depth priors by preserving smoothness and sharp boundaries, ultimately enhancing 3D reconstruction quality with similar computational efficiency.

\vspace{-20pt}

\begin{table}[htbp]
    \centering
        \caption{Quantitative results of SparseNeRF with depth regularization. SparseNeRF* is the original result report in \cite{sparsenerf}}
        \label{table:edgenerf_vs_sparsenerf}
        \begin{tabular}{c|ccc}
            \toprule
            & PSNR$\uparrow$ & SSIM$\uparrow$ & LPIPS$\downarrow$ \\
            \midrule
            SparseNeRF* & 19.86 & 0.624 &  0.328 \\
            SparseNeRF & \textbf{20.28} & 0.651 & 0.310 \\
            SparseNeRF + our depth regularization & 20.00 & \textbf{0.728} & \textbf{0.296} \\
            \bottomrule
        \end{tabular}
\end{table}

\vspace{-30pt}

\vspace{-20pt}

\section{Conclusion}

We present EdgeNeRF, an edge-guided approach for sparse views reconstruction, addressing the geometric ambiguity at object boundaries in existing global depth regularization methods. We apply adaptive regularization to depth and normal fields in non-edge regions at fine scales, while preserving natural discontinuities at detected edges. Extensive experiments demonstrate EdgeNeRF's superior performance and easy integration with other methods.

% \vspace{0.3\baselineskip}

\noindent\textbf{Limitations.} \ Current challenges include: (1) performance degradation with complex textures, (2) computational cost of normal regularization, and (3) semantic feature degradation from low-level smoothing (affecting LPIPS scores in DTU dataset). Future work may incorporate semantic-aware regularization.

\vspace{15pt}

\noindent\textbf{Acknowledgement.} \ This work was supported by the Basic and Applied
Basic Research Foundation of Guangdong Province
(2024A1515012287), Science and Technology Key Program of Guangzhou (2023B03J1388), National Key R\&D Program
of China (2023YFA1011601).

% \vspace{-5pt}

%
% ---- Bibliography ----
%
% BibTeX users should specify bibliography style 'splncs04'.
% References will then be sorted and formatted in the correct style.
%
\bibliographystyle{splncs04}
\bibliography{mybibliography}
%
% \begin{thebibliography}{8}
% \bibitem{ref_article1}
% Author, F.: Article title. Journal \textbf{2}(5), 99--110 (2016)

% \bibitem{ref_lncs1}
% Author, F., Author, S.: Title of a proceedings paper. In: Editor,
% F., Editor, S. (eds.) CONFERENCE 2016, LNCS, vol. 9999, pp. 1--13.
% Springer, Heidelberg (2016). \doi{10.10007/1234567890}

% \bibitem{ref_book1}
% Author, F., Author, S., Author, T.: Book title. 2nd edn. Publisher,
% Location (1999)

% \bibitem{ref_proc1}
% Author, A.-B.: Contribution title. In: 9th International Proceedings
% on Proceedings, pp. 1--2. Publisher, Location (2010)

% \bibitem{ref_url1}
% LNCS Homepage, \url{http://www.springer.com/lncs}. Last accessed 4
% Oct 2017
% \end{thebibliography}
\end{document}